# Authorship Detection of SMS Messages Using Unigrams

Roshan Ragel, Pramod Herath and Upul Senanayake
Department of Computer Engineering
Faculty of Engineering, University of Peradeniya
Peradeniya 20400 Sri Lanka

*Abstract*— SMS messaging is a popular media of communication. Because of its popularity and privacy, it could be used for many illegal purposes. Additionally, since they are part of the day to day life, SMSes can be used as evidence for many legal disputes. Since a cellular phone might be accessible to people close to the owner, it is important to establish the fact that the sender of the message is indeed the owner of the phone. For this purpose, the straight forward solutions seem to be the use of popular stylometric methods. However, in comparison with the data used for stylometry in the literature, SMSes have unusual characteristics making it hard or impossible to apply these methods in a conventional way. Our target is to come up with a method of authorship detection of SMS messages that could still give a usable accuracy. We argue that, considering the methods of author attribution, the best method that could be applied to SMS messages is an n-gram method. To prove our point, we checked two different methods of distribution comparison with varying number of training and testing data. We specifically try to compare how well our algorithms work under less amount of testing data and large number of candidate authors (which we believe to be the real world scenario) against controlled tests with less number of authors and selected SMSes with large number of words. To counter the lack of information in an SMS message, we propose the method of stacking together few SMSes.

## I. INTRODUCTION

Authorship detection is a research field that has been extensively explored in the last decade. This field is generally known as stylometry and had been in use long before computational techniques took a steep turn in its role. In the pre-in-silico era, authorship attribution was manually conducted by observing the linguistic information embedded in a text corpus. These techniques were based on linguistic markers such as frequency counts and word similarity [1]. Most of these efforts were focused on detection the authorship of literary texts. For instance, a lot of attention has been centred on attributing the authorship of historical text corpuses. One of the well-known studies in the field of authorship attribution is the case of Federalist papers in the USA where there was a dispute about twelve of the authors [2]. Another interesting investigation took place to identify the authorship of Shakespeare's plays in question [3].

Computational techniques used in authorship attribution follows a thorough pathway by managing to take many aspects into consideration as opposed to linguistic markers. These markers are taken into consideration using AI techniques such as statistical pattern recognition, machine learning and neural networks. Stylometry predominantly recognizes authorship attribution as recognizing the authorship of a non-adversarial author; however recently there has been much work carried out in adversarial stylometry as well which focuses on how to intentionally compose documents to avoid detection and how to detect these kinds of documents [4].

The main limitation of stylometric analysis is the size of the training corpus it needs to attribute an authorship. This is essentially not a limitation for compositions such as literary texts, articles or even emails. However this becomes a huge bottleneck when it comes to attributing authorship of compositions that are small such as SMS and Tweets. As such, different measures need to be taken in order to mitigate the limitations imposed by traditional stylometric techniques. The intention of the authors in this paper is to present a technique that can identify the authorship of SMS using the n-gram approach.

Unfortunately, like any other social tool, SMS messaging could be misused by malicious users to accomplish anti-social acts. These include spamming, harassments, distribution of misinformation and more serious issues such as communication between criminals. Since SMS data is not stored by the service providers, the original owners of the SMSes are not easily recognizable. In case a suspect is caught, authorship attribution could be used to verify whether it is the owner of the phone who has sent the malicious message.

The authors would like to point out that stylometric techniques are allowed as evidence in courts of the UK, the US and Australia [5]. SMS authorship as an evidence has been used mostly in the courts of the US where the case of Danielle Jones was an elaborative example [6]. The linguistic experts were able to analyse some SMSes which were supposedly sent by Danielle's mobile phone after she has been disappeared were actually not written by her but her uncle. Another well-known case study is the Jenny Nicholl case where linguistic experts have determined that the SMSes were authored by her ex-boyfriend [7]. Before continuing further, the authors wish to direct attention towards the following excerpt from the Institute for Linguistic Evidence in 2008.

"In some criminal, civil and security matters, language can be evidence. A suicide note, a threatening letter, anonymous communications, business emails, blog posts, trademarks – all of these can help investigators, attorneys, human resource executives and private individuals understand the heart of an incident. When you are faced with a suspicious document, whether you need to know who wrote it, or if it is a real threat or a real suicide note, or if it is too close for comfort to some other document, you need reliable validated methods." [8]

The intention of the authors is to establish reliable and validated method to analyse SMSes and determine their authorship in order to facilitate the materializing of legal evidence. A clear indication of the SMSes been used as evidence in the courts in the future is the recently passed Cybercrime Legislation Amendment Bill 2011 by the Australian Government which allows security forces to ask telecommunication companies to retain sensitive information which may potentially be used to identify threats. This sensitive information includes but not limited to SMSes or emails from terrorists, criminals, known drug dealers, etc. [9].

As such, authors would like to reiterate the importance of authorship attribution of SMSes as a security measure.

An SMS is a powerful way of communication. It is cheap and easy to use. Its asynchronous nature allows one to communicate without having to interact with the receiver directly. One can easily write anything on a message and send that he/she finds hard to say face-to-face. With these advantages, the usage of SMSes around the world has been increased exponentially in recent times.

Traditional stylometric analysis includes write invariants such as average sentence/word length usage frequencies of grammatical components such as nouns, verbs, adjectives, frequency of functional words and richness of vocabulary [10]. These methods fail in the case SMSes due to few reasons. The length of the SMS is limited to 140 characters. And since the size limitation of the cellular phone, many find it hard to type the words in. For these reasons the size of an average SMS is less and less consistent with grammar, spelling and punctuation. Additionally these cause the users to introduce shortened words for communicating through SMSes. There are standard popular SMS language words but mostly the shortened forms vary from user to user. In most locales where English is not the native language, users tend to transliterate the message in their local language, making it hard to automatically analyse for grammatical components. Even if you could device such methods, it is only applicable to that language. For different set of languages, much different software must be used.

Because of the problems discussed, the above mentioned methods mostly become inapplicable. For n-gram based methods on the other hand this scenario sets a good stage for categorization. Recurrent errors and unique shortened words actually make it easier for an n-gram based system to identify the real author. Since the sentence length is not taken in to consideration, punctuation will not cause problems in these methods. Since the method is language independent, it can be easily applied to any transliterated language. The problem of SMS is that it contains fewer amounts of data. When given a message to test for an author, it might not contain enough data to come into any conclusion with any method. To counter this, in our method we staked few SMS messages together to generate one testing dataset. Having a lot of SMSes from one unknown sender might not be the practical scenario. Therefore in our research we check for a lower boundary for an average lowest number of SMSes needed to have a considerable accuracy for identification of the author of an SMS.

When such a method is used, traditionally the stop words are dropped out in a pre-processing step to get a better result. However, in our case, having stop words actually helps the identification. Some SMS users do not use words like "a" or "the" since it makes the message longer while others still do use them. Therefore having that word in testing set could be an advantage towards correctly identifying the author of the message.

The rest of this paper is ordered as follows: section two consists of related work and section three discuss in brief on the subject of n-gram modelling. As the next section a short description of the distance metrics used by us for similarity check is presented. Next a brief introduction to NUS corpus is given. As the next section our methodology and results of author attribution is discussed. Next we conclude our paper with a discussion and a conclusion.

## II. RELATED WORK

The majority of research aims to attribute authorship of compositions that are sufficiently large such as literary texts or articles. There are a large amount of examples available for such experiments like the work carried out by Thisted and Efron [11], Pennebaker and King [12], Doddington [13], Wools [14], Slatcher et al. [15], Webber et al. [16], Shriberg and Stolcke [17], and Ishihara [18]. These are all significant developments in the field of stylometry.

A character n-gram based method of author attribution has been proposed by Keselj et al. [20]. In their method, the character n-grams are used to create a profile of an author. Character n-grams were used in this context since the aim of the authors was to come up with an algorithm that can be used across languages. Since in certain languages the word boundary is hard to decide, a word n-gram method fails. This is not a problem in our application since we assume that all SMSes are in Roman letters.

N-gram approach is used by Keselj et al. [21] to decide the gender. In this case author profiles were created using few different tokens such as characters, words, and part-of-speech tags. Unfortunately, this method is difficult to be applied to SMS messages since the nature of SMS makes it hard to come up with part of speech tagging due to previously mentioned challenges.

SMSes are much different from formal literature, and even from normal writing in paper. The difficulties in typing in cellular phones cause the messages to be short and unstructured. This difficulty also has caused for the introduction of a new shortened language for SMSes. Furthermore, since the SMSes are mostly informal, one tends to write in his/her own language and since the cellular phones usually support only English language, and even if other languages are available they are difficult to be typed, the messages tend to be transliterated, typically. All these facts make it hard to use syntactic means to decide the authorship of an SMS difficult.

On the other hand, lexical measures work well in this scenario. For an example, shortened words tend to be unique to each user. Which shortened word to use, how frequently it is used depends much on the personality of the author. We argue that a methodology such as unigram could be successful over syntactic analysis.

## III. N-GRAM MODELING

N-gram modelling is a very simple and powerful idea. N-gram models have been successfully applied in speech recognition [22], natural language processing [23] and spelling suggestion [24]. It is also successfully applied in author attribution [10].

An n-gram is 'n' number of consecutive sequences of tokens from a sequence of data. The tokens can be any unit of data in the data series. In our application specifically, the series of data is the string of text written by a certain author in an SMS while the token is a word. Alternatively, one could take characters as the token, making the model a character n-gram model. The letter 'N' represents the number of tokens taken together. In our application we considered only one token in the sequence, therefore called one-gram or unigram.

An n-gram model models the probability of an n-gram popping up next in the sequence and can be used as a lexical measurement for authorship attribution.

## IV. DISTANCE METRICS

In our experiments, we used both cosine similarity [25] and Euclidean distances [26] to check how well each of them perform in the task of measuring similarity of two vectors.

Cosine similarity is one method of measuring the distance between two vectors. It makes the use of the standard dot product of two vectors to find out the difference between two vectors.

Cosine distance θ is defined as,

$$similarity = \cos(\theta) = \frac{\sum_{i=1}^{n} A_i B_i}{\sqrt{\sum_{i=1}^{n} A_i} * \sqrt{\sum_{i=1}^{n} B_i}} \quad (1)$$

The other method we use in our experiment is the standard Euclidean distance metric. This is defined as follows:

$$Euclidean\ distance = \sum_{i=1}^{n} (A_i - B_i)^2 \quad (2)$$

## V. NUS CORPUS

NUS corpus [21] is one of the largest SMS message corpuses available. It contains more than fifty thousand messages written in English. The NUS corpus consists of messages that are from multiple cultures in Asia. Both transliterated and pure English data is available as NUS corpus has no restriction on the collected SMSes.

## VI. METHODOLOGY AND RESULTS

In our experiments, we have separated our SMS database into two groups: testing and training data. The training data set is used to create a profile of each author and was treated as data with a known author. We extracted the same profile from the testing data and this data was treated as data with author unknown. Taking the profiles of testing set of profiles one by one, we compared them with each of the training profiles belonging to each author. By checking the resulting similarity values, we chose the best author and labelled the testing SMS as belongs to that author. The accuracy of the algorithm was measured by the percentage of correct labelling.

As the author profile, we chose unigram counts. This is simply a collection of frequency of each word available in the string. We then made a vector from this data. For an example if the strings to be compared are: "this is a test string for test" and "this is a training string for training", the two profiles created could be the vectors: [1 1 1 2 1 1 0] and [1 1 1 0 1 1 2]. Note that each element in the two vectors represents the frequencies of the words "this", "is", "a", "test", "string", "for", and "training" respectively. Now the above two vectors can be compared with distance metrics presented in Section IV.

### A. Pre-processing

NUS database contains more than fifty thousand SMSes. We used the XML version of this database for our work. As a pre-processing step all SMSes from authors who have less than fifty collected SMSes were removed from the database. The database also included other types of messages such as multimedia messages and contact cards. These were also removed. It also contained duplicate messages which were also removed at the pre-processing stage.

### B. Choosing a suitable comparison method

The first experiment was carried out in order to determine a suitable comparison method. We had chosen two methods: cosine distance and Euclidean distance. For this experiment, about twenty authors were chosen having more than five hundred SMSes each. Out of these SMSes, five hundred longest SMSes were selected for the test.

For this experiment, a unigram authorship profiling was used. For all training data available, a distribution of unigrams was created for each author. Since the testing data in one SMS was not enough, we stacked few SMSes together to generate one large testing data. The extracted unigram profiles from the testing stack were then compared against the training set. For the comparison, the cosine similarity and the Euclidean distance between the training and the testing data were calculated separately. The author was then determined as the one having the highest similarity or the lowest distance.

Since the stacking of SMSes could have an effect on comparison, the number of SMSes stacked was varied. Starting from twenty SMSes stacked together to be tested against author profiles, it was varied down to one SMS in order to check the effects of the SMS stacking. The experiment was carried out with tenfold cross validation [23].

The result of this experiment clearly shows that Euclidean distances are not a good metric to measure the similarity of SMS author profiles. The difference is indicated in the graph depicted in Fig. 1.

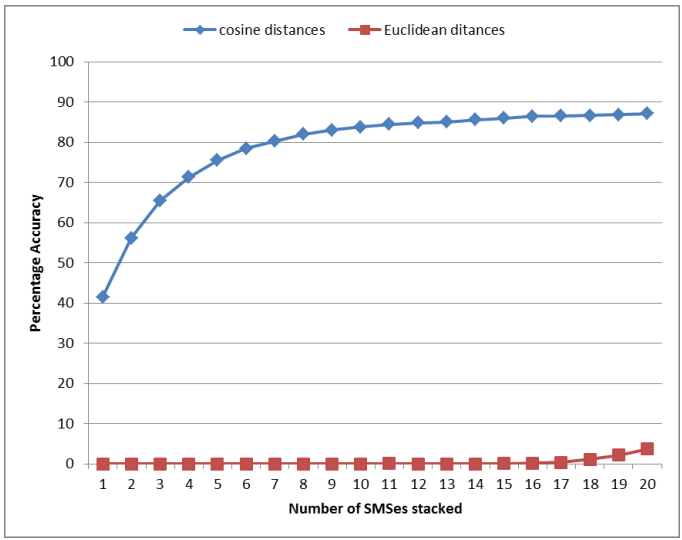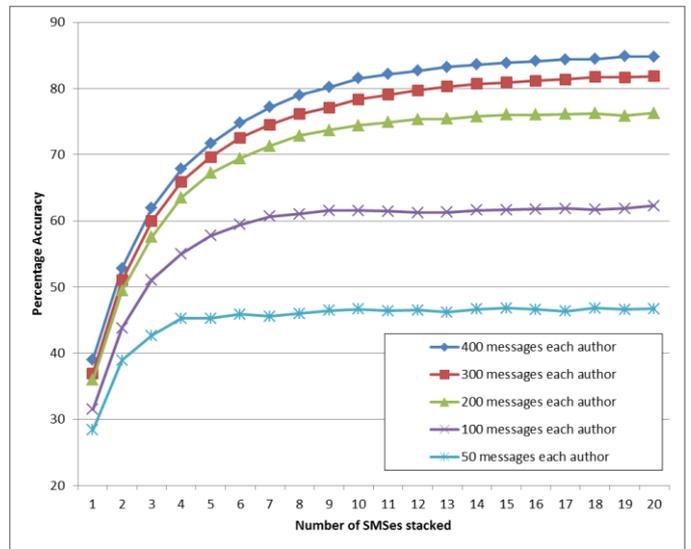

Fig. 1. Comparison between the results obtained by the cosine distance metric result and the Euclidean distance metric

Fig. 2. Effect of varying number of messages in the database

The graph in Fig. 1 shows that the Euclidean distances perform very poorly in comparison to the cosine distances. While the Cosine distances gives an accuracy starting from forty percent for only one SMS and going up to ninety percent when the number of SMSes stacked is large, Euclidean distance based method barely reaches five percent of accuracy.

The reason why this happens can be explained by considering how each metric treats missing data. The cosine similarity has no penalty over a missing data item in neither training nor testing data. But in Euclidean distances, a large penalty of squared the frequency of the available data is added. Our training corpus is a large one while the data in one testing instance is quite small. Therefore, there could be a huge penalty score for these missing data. Therefore, we conclude Cosine similarity is more suitable compared to Euclidean distance to measure similarity in our work. Therefore, for all our further experiments we chose cosine distance measurement as the metric to measure similarity.

### C. The effect of training data size

Our next experiment was carried out in order to determine the effect of the size of the training set (i.e. the database of SMSes whose authorship is known). The experimental setup was as same as in the previous experiment done to choose the suitable comparison methods except for the number of SMSes for training set was varied. We started from five hundred (the same experiment as before) and then started reducing the number of training SMSes in a step of hundred SMSes at a time. As same as before, we changed the number of testing SMSes from one to twenty in order to determine its effect. The experimental results were obtained with tenfold cross validation. Fig. 2 shows the results we obtained.

The graph in Fig. 2 shows that, as the number of test SMSes stacked together increase the accuracy of the results will increase. However, still it reaches a saturation level around twenty SMSes for testing. In addition, a difference between hundred SMSes has not introduced a considerably higher accuracy when 300 SMSes per author and 400 SMSes per author instances are compared.

This result clearly shows how successful the simple unigram method is on author attribution of SMSes. Though stacking twenty SMSes together gives eighty per cent accuracy for four hundred collected SMSes from each author, even stacking ten SMSes gives a closer value.

### D. The effect of training data set and reality

In a real world situation, neither testing nor training set will be of maximum length. In our previous two experiments, we deliberately chose the maximum length SMSes. Practically, this might not be possible because of the short comings of the database. Here we try to simulate such a situation by randomly selecting the SMSes without selecting the longest SMSes available.

The experimental setup was as of the last experiment, except that we implemented a pseudo-random number generator to pick up the number of SMSes needed from each author.

This result in Fig. 3 clearly shows how successful is the algorithm on randomly selected data. This is closer to the real world scenario than the previous one and still shows a close rate of accuracy. Note that the saturation characteristic observed in the previous experiment is still observable here.

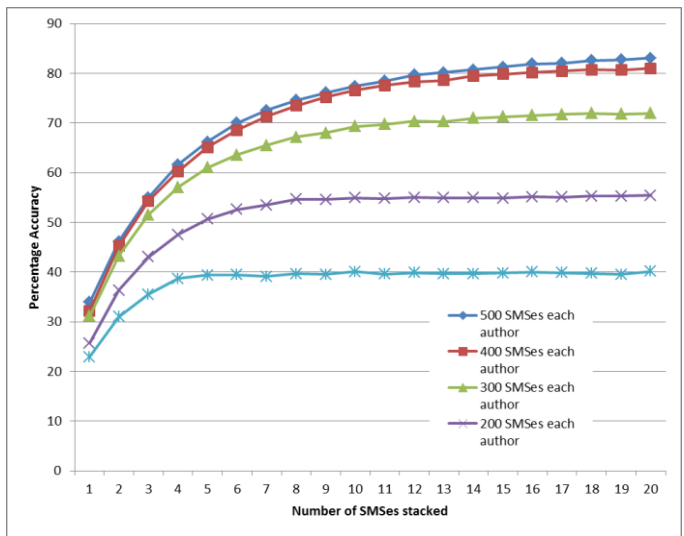

Fig. 3. Random selection of SMSes as training set

The graph in Fig. 3 is very similar to the graph in Fig. 2 where the experiment is carried out with a carefully selected data set. It can be concluded that careful choice of long SMSes has not really affected the results.

*E. The effect of number of authors in the dataset*

The algorithm might be successful in attributing the correct author in a limited scenario where only few candidate authors are there to choose from. But in a real world scenario there might be many more suspects to choose the author from. The next experiment is designed to check how well it works for varying number of authors with little data available. The experiment starts with five candidate authors and goes up to seventy authors. Each one has fifty SMSes as the training set. To choose the authors a random number generator is used.

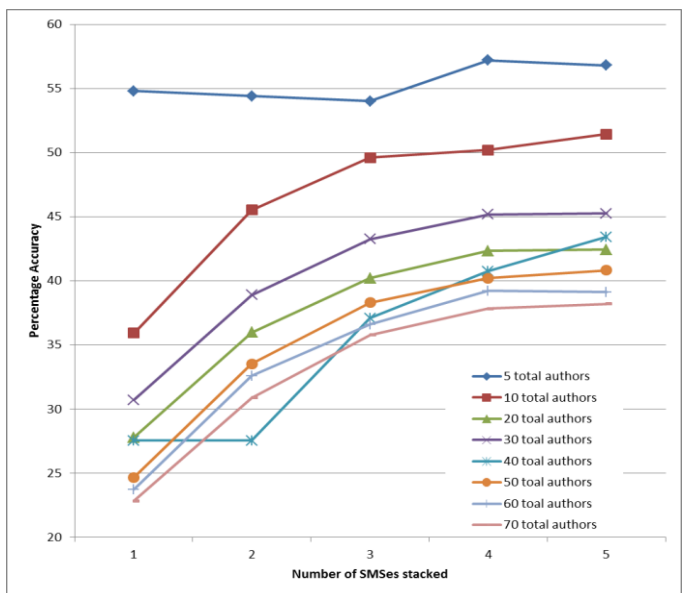

Fig. 4. The effect of varying the number of possible authors

From Fig. 4 it can be observed that when the number of possible authors increases, the decrease in the accuracy is close to linear.

Fig. 4 shows how the result varies with the number of possible authors. We specifically chose lower number of testing SMSes to note how the algorithm handles lower number of data. For this experiment, only fifty training sets were chosen. We observe that there is around thirty per cent reduction of accuracy when the number of possible authors increases from five to seventy.

## VII. DISCUSSION

Following observations are made in the results that have been shown in Section VI:

Even with very small amount of testing data (even with one SMS to test) the algorithm produces good results. But the growth of accuracy as the number of testing SMSes grows is not linear. The accuracy growth rate goes down with the increase of testing data until it saturates around ninety per cent. But a good accuracy has been achieved with a small number of test cases (around ten SMSes stacked together).

Euclidean distance is not a very good method of comparing two author profiles in comparison to cosine similarity method. This might be due to the sparse data problem. Euclidean distance gives a large penalty (i.e., frequency squared) if a data point is missing. This characteristic is not shown in cosine similarity. Therefore for this application without smoothing, cosine similarity is better.

## VIII. CONCLUSION

In our experiments, we tried to identify the best conditions for author attribution of SMS messages with the unigram method. According to our findings, around ten SMSes stacked together could produce a good test data to detect the author of the set of SMSes with the near best accuracy possible for that dataset. Cosine similarity would be a good choice to attribute the authors.